\def\paperlanguage{}
\pdfoutput=1 

\documentclass[letterpaper, 10 pt, conference]{ieeeconf}  

\usepackage{bm}
\usepackage{cite}
\usepackage{flushend}
\include{preamble}

\IEEEoverridecommandlockouts                              

\title{\LARGE \textbf
  {
    \switchlanguage%
    {%
      Design Optimization of Musculoskeletal Humanoids with Maximization of Redundancy to Compensate for Muscle Rupture
    }%
    {%
      筋破断を補償する冗長性を最大限活用した筋骨格ヒューマノイドの設計最適化
    }%
  }
}

\author{Kento Kawaharazuka$^{1}$, Yasunori Toshimitsu$^{1}$, Manabu Nishiura$^{1}$, Yuya Koga$^{1}$,\\Yusuke Omura$^{1}$, Yuki Asano$^{1}$, Kei Okada$^{1}$, Koji Kawasaki$^{2}$, and Masayuki Inaba$^{1}$
  \thanks{$^{1}$ The authors are with the Department of Mechano-Informatics, Graduate School of Information Science and Technology, The University of Tokyo, 7-3-1 Hongo, Bunkyo-ku, Tokyo, 113-8656, Japan.
    {\texttt\small [kawaharazuka, toshimitsu, nishiura, koga, oomura, asano, k-okada, inaba]@jsk.t.u-tokyo.ac.jp}
  }
  \thanks{$^{2}$ The author is associated with TOYOTA MOTOR CORPORATION.
    {\texttt\small koji\_kawasaki@mail.toyota.co.jp}
  }
}
\begin{document}

\maketitle
\thispagestyle{empty}
\pagestyle{empty}

\begin{abstract}
  \switchlanguage%
  {%
    Musculoskeletal humanoids have various biomimetic advantages, and the redundant muscle arrangement allowing for variable stiffness control is one of the most important.
    In this study, we focus on one feature of the redundancy, which enables the humanoid to keep moving even if one of its muscles breaks, an advantage that has not been dealt with in many studies.
    In order to make the most of this advantage, the design of muscle arrangement is optimized by considering the maximization of minimum available torque that can be exerted when one muscle breaks.
    This method is applied to the elbow of a musculoskeletal humanoid Musashi with simulations, the design policy is extracted from the optimization results, and its effectiveness is confirmed with the actual robot.
  }%
  {%
    筋骨格ヒューマノイドは生物模倣型の様々な利点を持つが, その中でも可変剛性制御を可能とする冗長な筋配置は最も重要な利点の一つである.
    本研究では, これまで多く扱われてこなかった, 冗長な筋配置により筋が一本切れても動き続けられるという利点を扱う.
    この利点を最大限利用するため, 筋が一本切れた際に発揮できるトルクの最小値最大化を考え, 設計の最適化を行う.
    本手法をシミュレーションと筋骨格ヒューマノイドMusashiの肘に適用し, 最適化結果からの設計方針の抽出とその有効性を確認する.
  }%
\end{abstract}

\section{Introduction}\label{sec:introduction}
\begin{figure}[t]
  \centering
  \includegraphics[width=1.0\columnwidth]{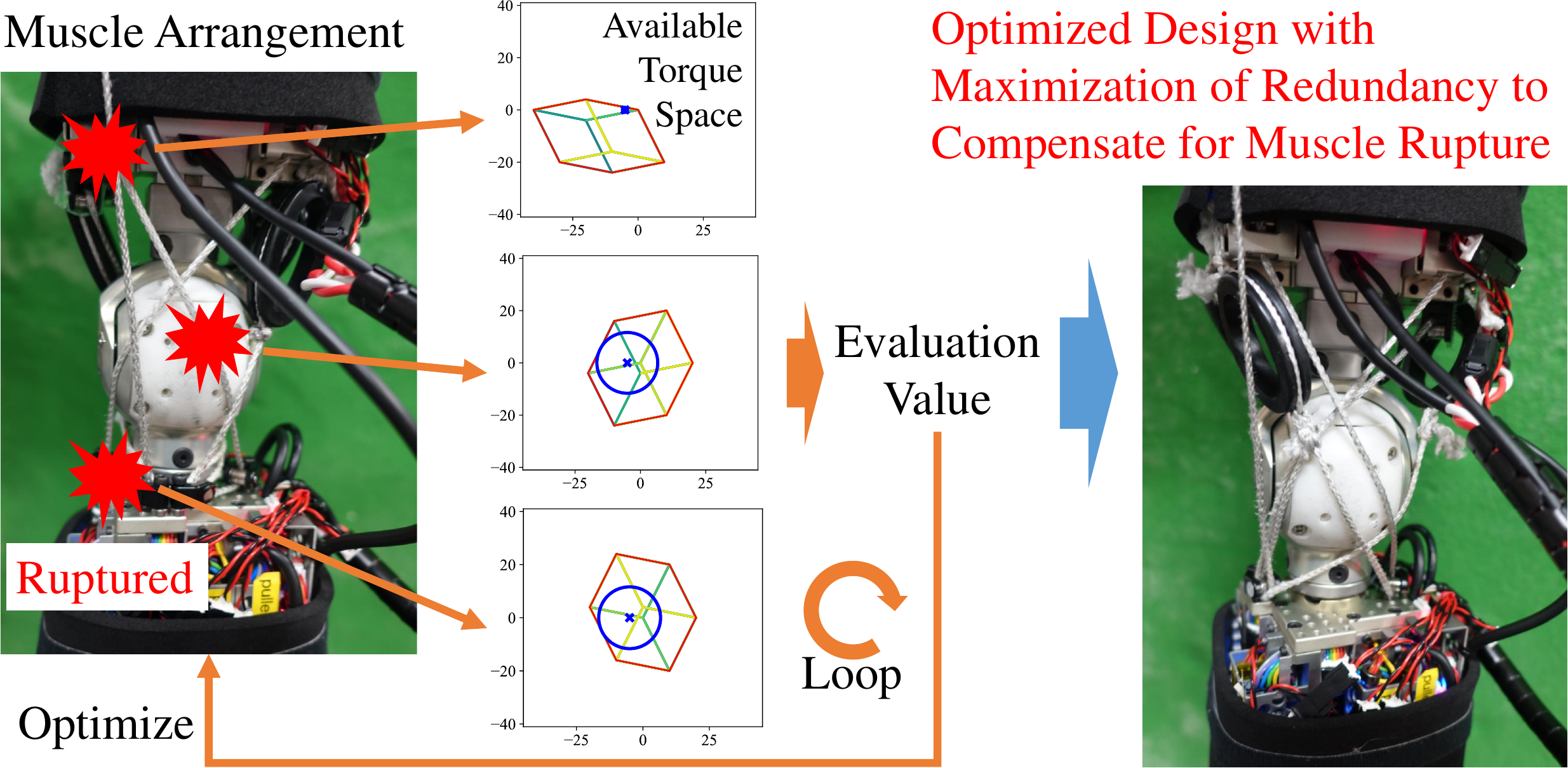}
  \vspace{-3.0ex}
  \caption{The concept of this study. By calculating the radius of the hypersphere (blue circle) inscribed in the available torque space (red polygon) when each muscle is broken, an evaluation value is calculated, and the design (muscle Jacobian) is optimized by a genetic algorithm so that the robot can continue to move even when one muscle is broken.}
  \label{figure:concept}
  \vspace{-3.0ex}
\end{figure}

\switchlanguage%
{%
  The musculoskeletal humanoid \cite{nakanishi2013design, wittmeier2013toward, kawaharazuka2019musashi, jantsch2013anthrob} has various biomimetic advantages such as ball joints without singular points, the flexible spine and fingers, the wide range motion of the scapula, and redundant muscle arrangements allowing for variable stiffness control.
  Among them, the redundant muscle arrangement is one of its most characteristic structures, with various advantages and disadvantages.
  So far, using its advantages, variable stiffness control with nonlinear elastic elements \cite{kobayashi1998tendon, kawaharazuka2019longtime}, balancing and joint coordination with biarticular muscles as well as monoarticular muscles \cite{sharbafi2016biarticular}, maximization of end-point force and joint angle velocity with redundant muscles \cite{marjaninejad2019redundant}, etc. have been studied.
  Also, methods to reduce high muscle tension due to antagonism \cite{kawaharazuka2017antagonist, koga2019modification}, methods to solve the problem of joint angle speed limited by the slowest muscle \cite{kawaharazuka2020speed}, etc. have been developed to compensate for its disadvantages.

  In this study, we focus on one feature of this redundant muscle arrangement, that is, the advantage of being able to continue to move using the redundant muscles even if one muscle is broken.
  This feature has not been widely discussed in the past.
  As related works, a method for calculating muscle tension without using a broken muscle in muscle tension-based control \cite{kawamura2016jointspace} and a method to keep moving even when a muscle breaks through online learning in muscle length-based control \cite{kawaharazuka2019longtime} have been developed.
  In humans, the change in the feasible force set when some muscles are paralyzed has been discussed \cite{kuxhaus2005palsy}.
  However, although the analysis of whether the human body can move as expected when the muscle breaks has been conducted, musculoskeletal humanoids have not yet successfully achieved the performance that allows it to keep moving in all directions even when one muscle is broken.

  Therefore, the purpose of this study is a development of body design optimization method for maximizing the redundancy to compensate for muscle rupture.
  The concept is shown in \figref{figure:concept}, which evaluates how the available torque space changes when the muscle breaks and runs an optimization loop.
  In \secref{sec:musculoskeletal-humanoids}, we describe the basic structure of musculoskeletal humanoids.
  In \secref{sec:proposed-method}, we describe the calculation of an index for evaluating the redundancy and a design optimization method by maximizing the index.
  In \secref{sec:experiments}, we analyze the optimization results of the 1-DOF (degree of freedom) and 2-DOF joint mechanisms in simulation, and apply them to the musculoskeletal humanoid Musashi \cite{kawaharazuka2019musashi}.
  Lastly, the results of the experiments are discussed in \secref{sec:discussion} and the conclusions are presented in \secref{sec:conclusion}.
}%
{%
  筋骨格ヒューマノイド\cite{nakanishi2013design, wittmeier2013toward, jantsch2013anthrob, kawaharazuka2019musashi}には, 特異点のない球関節, 柔軟な背骨や指, 可動域の広い肩甲骨, 可変剛性制御を可能とする冗長な筋配置等, 様々な生物模倣型の利点を有する.
  この中でも, 冗長な筋配置は様々な利点・欠点を持つ特徴的な構造である.
  これまで, その利点を活かし, 非線形性要素と冗長性を利用した可変剛性制御\cite{kobayashi1998tendon, kawaharazuka2019longtime}, 1関節筋だけでなく2関節筋を用いたバランシングと関節協調\cite{sharbafi2016biarticular}等が行われている.
  また, 拮抗による高張力を削減する方法\cite{kawaharazuka2017antagonist, koga2019modification}, 最も遅い筋により関節速度が制限される問題を解決する方法\cite{kawaharazuka2020speed}等, その欠点を補う手法も様々開発されている.

  本研究では, この冗長な筋配置の一つの特徴である, 筋が一本切れても冗長な筋肉を使って動き続けられるという利点に焦点を当てる.
  この特徴はこれまで多くは議論されていない.
  関連研究としては, 関節トルク制御において切れた筋を使わずに筋張力を算出する方法\cite{kawamura2016jointspace}, 筋長制御においてオンライン学習により筋が切れても動き続ける方法\cite{kawaharazuka2019longtime}が開発されている.
  人間においても, 一部の筋が麻痺した場合におけるfeasible force setの変化についての考察が行われている.
  しかし, これまでは筋が切れた時に, そもそも身体を思ったように動かせるかどうかを判断することや, その指標を元に設計を最適化すること等は行われていない.

  そこで本研究では, 筋破断を補償する冗長性を最大限に利用するための設計方針を最適化によって明らかにする.
  そのコンセプトは\figref{figure:concept}に示す通りであり, 筋が切れたときにどう発揮可能関節トルク空間が変化するかどうかを評価し, 最適化のループを回す.
  \secref{sec:musculoskeletal-humanoids}では筋骨格ヒューマノイドの基本的な筋構造について説明する.
  \secref{sec:proposed-method}では冗長性を評価するための評価値計算, その最大化による設計最適化について説明する.
  \secref{sec:experiments}ではシミュレーションにおいて1DOF, 2DOFの関節機構の解析, その筋骨格ヒューマノイドMusashi \cite{kawaharazuka2019musashi}への適用について述べる.
  最後に, \secref{sec:discussion}でそれら実験結果について議論し, \secref{sec:conclusion}で結論を述べる.
}%

\section{Musculoskeletal Humanoids} \label{sec:musculoskeletal-humanoids}
\begin{figure}[t]
  \centering
  \includegraphics[width=0.6\columnwidth]{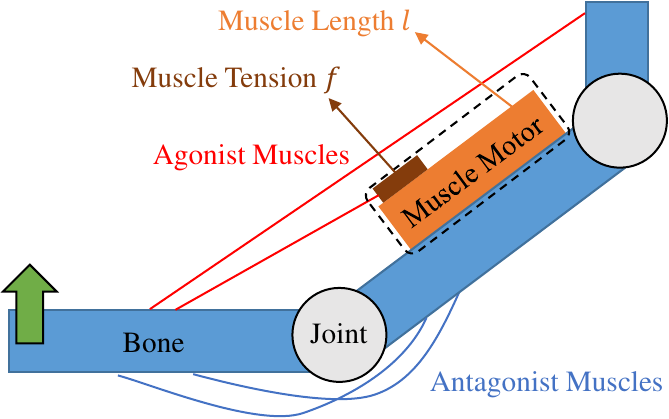}
  \vspace{-1.0ex}
  \caption{The basic musculoskeletal structure.}
  \label{figure:basic-structure}
  \vspace{-1.0ex}
\end{figure}

\switchlanguage%
{%
  The basic structure of the musculoskeletal humanoid is shown in \figref{figure:basic-structure}.
  The muscles are redundantly arranged around the joint.
  Muscles that move the joint in the direction of the intended movement are called ``agonist muscles'' and those that move in the direction that prevents the movement are called ``antagonist muscles''.
  Muscle tension $f$ and muscle length $l$ are usually measured.
  The joint angle cannot usually be measured due to the complexity of the scapula and ball joints, but some robots have joint angle sensors with special mechanisms for learning controls or experimental evaluation.

  There are several types of musculoskeletal humanoids.
  First, there are a few types of actuators, such as those with electric motors and pulleys winding a wire \cite{nakanishi2013design}, those with pneumatic muscles \cite{niiyama2010design}, and those with twisted and coiled polymer (TCP) \cite{almubarak2017twisted}.
  Second, there are two types of muscle arrangements: one in which the moment arm of muscles is constant and easy to modelize, and the other in which the moment arm changes like a human being, which is more complicated and difficult to modelize.
  The musculoskeletal humanoid used in this study uses electric motors and has a complex musculoskeletal structure with variable moment arm like human beings.
  However, in the case where the moment arm changes, it is difficult to analyze the muscle arrangement numerically.
  Therefore, we assume that the moment arm is constant to some extent as long as the joint does not move too largely, and analyze the muscle arrangement as if the moment arm is constant.

  As for the musculoskeletal humanoid Musashi \cite{kawaharazuka2019musashi} used in this study, the body is constructed by using muscle modules including motors, gears, pulleys to wind the muscles, muscle tension measurement units, motor drivers, and temperature sensors.
  The direction in which the muscle exits from the motor module can be arbitrarily changed, and by using three types of muscle relay units, various muscle paths and moment arms can be realized.
  The muscle wire is made of Dyneema\textsuperscript{\textregistered}, an abrasion resistant synthetic fiber.
  The gear ratios of the motors used in this study are all 29:1, and they are backdrivable.
  In this study, we will consider the problem of determining the maximum value of the moment arm that can be realized, optimizing the muscle Jacobian within that range, and realizing it by arranging the muscle modules and muscle relay units.
}%
{%
  筋骨格ヒューマノイドの基本的な構造を\figref{figure:basic-structure}に示す.
  関節の周りに筋が冗長に配置されている.
  意図した動作を行う方向に関節を動かす筋を主動筋, その動きを妨げる方向に関節を動かす筋を拮抗筋と呼ぶ.
  筋ワイヤは摩擦に強い化学繊維であるDyneemaを用いており, 筋張力$f$, 筋長$l$を測定できる.
  関節角度は複雑な肩甲骨や球関節のため基本的には測定できないが, 学習や実験評価のために存在する場合もある.

  筋骨格ヒューマノイドにも種類が存在する.
  アクチュエータについては, 電気モータとプーリによりワイヤを巻き取る方式\cite{nakanishi2013design}, 空気圧による人工筋肉を用いる方式\cite{niiyama2010design}, Twisted and coiled polymer (TCP)を用いる方式\cite{almubarak2017twisted}等が存在する.
  また, 筋のモーメントアームが一定であるモデル化が容易な方式, 人間のようにモーメントアームが変化する, より複雑でモデル化が困難な方式が存在する.
  本研究の実機で扱うロボットは, この中でも電気モータを用いており, 人間のようにモーメントアームが変化する複雑な筋骨格構造を持つ.
  しかし, モーメントアームが変化する場合はモデル化が困難なため, 数値的な解析が難しい.
  そのため, 本研究では関節角度限界範囲の中心から大きく動かなければモーメントアームはある程度一定と捉えることができると考え, モーメントアーム一定として解析を行う.
}%

\begin{figure}[t]
  \centering
  \includegraphics[width=0.6\columnwidth]{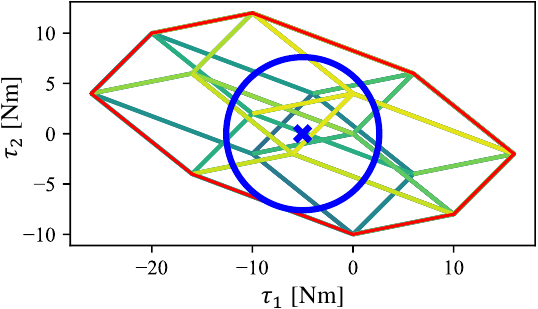}
  \vspace{-1.0ex}
  \caption{Hypersphere (blue circle) inscribed in available torque space (red polygon). The line with a gradient from green to yellow is the line where each edge of the available muscle tension space is transformed into the joint torque space.}
  \label{figure:maximum-circle}
  \vspace{-3.0ex}
\end{figure}

\section{Design Optimization with Maximization of Redundancy to Compensate for Muscle Rupture} \label{sec:proposed-method}
\subsection{Calculation of RITS}
\switchlanguage%
{%
  In this section, we derive an index required to evaluate the redundancy.
  The basic equations for the musculoskeletal structure are given below,
  \begin{align}
    \bm{l} &= \bm{h}(\bm{\theta})\\
    d\bm{l} &= Gd\bm{\theta}\\
    \bm{\tau} &= -G^T\bm{f} \label{eq:convert-tension}
  \end{align}
  where $\bm{l}$ is the muscle length, $\bm{f}$ is the muscle tension, $\bm{\theta}$ is the joint angle, $\bm{\tau}$ is the joint torque, $\bm{h}$ is the mapping from $\bm{\theta}$ to $\bm{l}$, and $G$ is the muscle Jacobian.
  Also, $\{\bm{l}, \bm{f}\}$ are $M$-dimensional vectors ($M$ is the number of muscles) and $\{\bm{\theta}, \bm{\tau}\}$ are $N$-dimensional vectors ($N$ is the number of joints).
  In the case with the variable moment arm, $G$ can be expressed as $G(\bm{\theta})$, and in the case with the constant moment arm, $G$ is a constant matrix (this study).

  The purpose of this section is to obtain the Radius of hypersphere Inscribed in available Torque Space (RITS), which is equivalent to the radius of the blue circle shown in \figref{figure:maximum-circle} if the torque space is 2-dimensional.
  The red line represents the boundary of the available torque space.
  This is the same idea as the Maximum Isotropic Value (MIV) used in \cite{finotello1998computation, inouye2014anthropomorphic}, but differs in terms of thinking about joint torque and determining whether the origin exists in the available torque space.
  Also, basic tendon-driven mechanisms and discussions of feasible force set are detailed in \cite{valero2015neuromechanics}.
  Since RITS is an index of how much torque can be exerted in all joint directions, the robot can move the body if the value of RITS is greater than zero, even if one muscle is broken.
  RITS $r$ can be calculated as follows,
  \begin{align}
    \textrm{maximize}\;\;\;\;\;\;&\;\;\;\;\;r\\
    \textrm{subject to}\;\;\;\;\;&R \subseteq T\\
    & R := \{\bm{\tau} \in \mathbb{R}^{N} \mid |\bm{\tau}-\bm{\tau}_{g}| \leq r\}\\
    & T := \{\bm{\tau} \in \mathbb{R}^{N} \mid \exists\bm{f} \in F, \bm{\tau} = -G^T\bm{f}\} \label{eq:torque-space}\\
    & F := \{\bm{f} \in \mathbb{R}^{M} \mid \bm{f}^{min} \leq \bm{f} \leq \bm{f}^{max}\} \label{eq:tension-space}
  \end{align}
  where $\bm{f}^{\{min, max\}}$ is the minimum or maximum value of muscle tension, and $\bm{\tau}_{g}$ is the joint torque that must be generated against gravity, friction, or external force.
  In this study, the $\bm{\tau}_g$ is assumed to be constant.

  The method to obtain this $r$ is explained below.
  First, in \equref{eq:tension-space}, the range of $\bm{f}$ can be taken as a $M$-dimensional hypercube $F$.
  Then, we can consider the $F$ to be converted into a $N$-dimensional hyperpolyhedron $T$ by \equref{eq:convert-tension}.
  Here, even if a convex hyperpolyhedron is converted to a hyperpolyhedron by linear transformation, it is still convex.
  Therefore, we can consider $r$ to be the radius of the hypersphere inscribed in this $N$-dimensional hyperpolyhedron $T$.
  The calculation algorithm can be written as \algoref{algorithm:calc-hoge}.

  \begin{algorithm}[t]
    \caption{Calculation of RITS $\bm{r}$}
    \label{algorithm:calc-hoge}
    \begin{algorithmic}[1]
      \Function{CalcRITS}{$G, \bm{\tau}_{g}$}
        \State $included \gets $True
        \State $\bm{p} \gets []$
        \For{$\bm{v}_F$ in all vertices of $F$}
          \State push $-G^T\bm{v}_F$ to $\bm{p}$
        \EndFor
        \State $C \gets \textrm{CalcConvexHull}(\bm{p})$
        \State $\bm{c} \gets \textrm{CalcCenter}(C)$
        \State $r \gets 1e9$
        \For{$\bm{s}_C$ in all hyperplanes of $C$}
          \State $d_1 \gets \textrm{CalcDistanceWithSign}(\bm{\tau}_{g}, \bm{s}_C)$
          \State $d_2 \gets \textrm{CalcDistanceWithSign}(\bm{c}, \bm{s}_C)$
          \If{$d_1d_2 < 0$}
            \State $included \gets $ False
          \EndIf
          \State $r \gets \min(|d_1|, r)$
        \EndFor
        \If{not $included$}
          \State $r \gets 0$
        \EndIf
        \State \Return $\bm{r}$
      \EndFunction
    \end{algorithmic}
  \end{algorithm}

  First, all the vertices of the hypercube $F$ are taken out and each of them are projected into the space $T$ of $\bm{\tau}$.
  Then, we calculate the convex hull $C$ of all projected vertices ($\textrm{CalcConvexHull}$).
  If this $C$ does not contain the vertex $\bm{\tau}_{g}$, then the inscribed hypersphere does not exist and we need to remove it.
  First, we calculate the center point $\bm{c}$ that is the average of all vertices in $C$ that definitely exists in $C$ ($\textrm{CalcCenter}$).
  Then, for all the hyperplanes in $C$, we calculate the signed distance $d_1$ from $\bm{\tau}_{g}$ and that of $d_2$ from $\bm{c}$ ($\textrm{CalcDistanceWithSign}$).
  This is equivalent to finding the normal $\bm{a}$ of the hyperplane and calculating $(\bm{a}^{T}\bm{\tau}_{g}-1)/||\bm{a}||$ or $(\bm{a}^{T}\bm{c}-1)/||\bm{a}||$.
  If the product of these two signed distances, $d_{1}d_{2}$, is positive, then $\bm{c}$ and $\bm{\tau}_{g}$ are on the same side of the hyperplane.
  If $\bm{\tau}_{g}$ is on the same side of the hyperplane as $\bm{c}$ with respect to all the hyperplanes in $C$, then $\bm{\tau}_{g}$ exists in $C$.
  If the inscribed hypersphere exists, we calculate the distance $|d_1|$ from $\bm{\tau}_{g}$ to the hyperplane in $C$, and its minimum value is RITS.
}%
{%
  本節では冗長性の評価値を計算するために必要な値(RITS)の導出を行う.
  筋骨格構造における基本的な式を以下に示す.
  \begin{align}
    \bm{l} &= \bm{h}(\bm{\theta})\\
    d\bm{l} &= Gd\bm{\theta}\\
    \bm{\tau} &= -G^T\bm{f} \label{eq:convert-tension}
  \end{align}
  ここで, $\bm{l}$は筋長, $\bm{f}$は筋張力, $\bm{\theta}$は関節角度, $\bm{\tau}$は関節トルク, $\bm{h}$は関節角度から筋長へのマッピング, $G$は筋長ヤコビアンを表す.
  また, $\{\bm{l}, \bm{f}\}$は$M$次元のベクトル($M$は筋の数を表す), $\{\bm{\theta}, \bm{\tau}\}$は$N$次元のベクトル($N$は関節の自由度数を表す)である.
  モーメントアームが一定でない場合, $G$は$G(\bm{\theta})$となり, 本研究のように一定と仮定する場合は定数行列となる.

  ここで本節の目的は, 発揮可能関節トルク空間に内接する超球の半径(Radius of hypersphere Inscribed in available Torque Space, RITS)を求めることである.
  これは, 例えばトルク空間を2次元とすると, \figref{figure:maximum-circle}に示す青い円の半径と同等である.
  赤い線は発揮可能トルク空間の境界を表す.
  これは\cite{finotello1998computation, inouye2014anthropomorphic}で用いられているMIV(Maximum Isotropic Value)と同様の考え方であるが, 本研究ではこれを関節トルクについて考える点や原点が発揮可能関節トルク空間内に存在するかどうかを判定する点等で異なる.
  また, 発揮可能トルク空間のsizeやshapeに関する議論は\cite{valero2015neuromechanics}に詳しい.
  RITSは全関節方向に対してどの程度力を発揮することができるかの指標となるため, 筋が切れた状態でも, この値が0よりも大きければ身体を動かすことが可能となる.
  RITS $r$は数式で表すと以下のように計算される.
  \begin{align}
    \textrm{maximize}\;\;\;\;\;\;&\;\;\;\;\;r\\
    \textrm{subject to}\;\;\;\;\;&R \subseteq T\\
    & R := \{\bm{\tau} \in \mathbb{R}^{N} \mid |\bm{\tau}-\bm{\tau}_{g}| \leq r\}\\
    & T := \{\bm{\tau} \in \mathbb{R}^{N} \mid \exists\bm{f} \in F, \bm{\tau} = -G^T\bm{f}\} \label{eq:torque-space}\\
    & F := \{\bm{f} \in \mathbb{R}^{M} \mid \bm{f}^{min} \leq \bm{f} \leq \bm{f}^{max}\} \label{eq:tension-space}
  \end{align}
  ここで, $\bm{f}^{\{min, max\}}$は筋張力の最小・最大値, $\bm{\tau}_{g}$は重力や摩擦等に抗って出さなければならない関節トルクを表す.
  本研究では, この$\bm{\tau}_g$は一定とみなしている.

  この$r$を求める方法を以下に示す.
  まず, \equref{eq:tension-space}において, この$\bm{f}$は$M$次元の超立方体$F$と捉えることができる.
  そして, この$\bm{f}$が\equref{eq:convert-tension}によって, $N$次元の$\bm{\tau}$の超多面体$T$へと変換されると考えることができる.
  ここで, 凸な超多面体を一次変換により超多面体に変換してもこれは凸である.
  そのため, この$r$は, この$N$次元超多面体内に内接する超球の半径と考えることができる.
  その計算アルゴリズムは\algoref{algorithm:calc-hoge}のように書ける.

  \begin{algorithm}[t]
    \caption{Calculation of RITS $\bm{r}$}
    \label{algorithm:calc-hoge}
    \begin{algorithmic}[1]
      \Function{CalcRITS}{$G, \bm{\tau}_{g}$}
        \State $included = $True
        \State $\bm{p} = []$
        \For{$\bm{v}_F$ in all vertices of $F$}
          \State push $-G^T\bm{v}_F$ to $\bm{p}$
        \EndFor
        \State $C = \textrm{CalcConvexHull}(\bm{p})$
        \State $\bm{c} = \textrm{CalcCenter}(C)$
        \State $r = 1e9$
        \For{$\bm{s}_C$ in all simplices of $C$}
          \State $d_1 = \textrm{CalcDistanceWithSign}(\bm{\tau}_{g}, \bm{s}_C)$
          \State $d_2 = \textrm{CalcDistanceWithSign}(\bm{c}, \bm{s}_C)$
          \If{$d_1d_2 < 0$}
            \State $included = $ False
          \EndIf
          \State $r = \min(|d_1|, r)$
        \EndFor
        \If{not $included$}
          \State $r = 0$
        \EndIf
        \State \Return $\bm{r}$
      \EndFunction
    \end{algorithmic}
  \end{algorithm}

  まず, $\bm{f}$が張る超立方体$F$の全頂点を取り出し, それを$\bm{\tau}$の空間$T$へと射影する.
  そして, その射影された全長点のconvex hull $C$を計算する($\textrm{CalcConvexHull}$).
  この$C$内に頂点$\bm{\tau}_{g}$が含まれなければ内接超球は存在しないため, これを判定する必要がある.
  まず, 絶対に$C$内に存在する, $C$に含まれる全頂点の平均である中心点$\bm{c}$を求める($\textrm{CalcCenter}$).
  次に, $C$内の全超平面について,$\bm{\tau}_{g}$からの距離と$\bm{c}$からの距離を符号付きで求める($\textrm{CalcDistanceWithSign}$).
  これは, 超平面の法線$\bm{a}$を求め, $(\bm{a}^{T}\bm{\tau}_{g}-1)/||\bm{a}||$を計算することに値する.
  この二つの距離$d_1, d_2$の掛け算が正であるとき, この$\bm{c}$と$\bm{\tau}_{g}$は超平面に対して同じ側に存在する.
  これを全超平面に対して行い, 全て$\bm{c}$と同じ側にあった場合, $C$内に$\bm{\tau}_{g}$が存在することになる.
  存在した場合, $\bm{\tau}_{g}$から$C$によって得られた超平面への距離$|d_1|$を求め, この最小値が内接する超球の半径となる.
}%

\subsection{Evaluation of Redundancy}
\switchlanguage%
{%
  For a given design ($G$), we calculate an index representing whether or not the robot with the design continues to move even if one muscle breaks, using RITS.
  Let $G_i$ be the $G$ where the moment arm of the $i$-th row of $G$ becomes zero, i.e. the $i$-th muscle is broken.
  In this study, we use the following $E$ as the evaluation value,
  \begin{align}
    r_0 &= \textrm{CalcRITS}(G, \bm{\tau}_{g})\\
    r_i &= \textrm{CalcRITS}(G_i, \bm{\tau}_{g})\;\;\;(1 \leq i \leq M)\\
    E_{value} &= \Sigma^{M}_{i=0}{r_i}\\
    E_{count} &= \Sigma^{M}_{i=0}{\textrm{Integer}(r_i > 0)}\\
    E &= \begin{cases}
      E_{value}\;\;\;(E_{count} \geq M_{min}+1)\\
      0\;\;\;\;\;\;(otherwise)
    \end{cases}
  \end{align}
  where $M_{min}$ is the lower limit of the number of muscles that can possibly be broken (only one of these muscles can be broken at a time), and $\textrm{Integer}(x)$ is a function that returns 1 if $x$ is true and 0 if false.
  That is, $E_{count}-1$ is the number of muscles that can possibly be broken, and if $E_{count}-1 \geq M_{min}$, the total RITS value for $G$ and $G_i$ for each muscle rupture is $E$.
}%
{%
  与えられた設計($G$)について, RITSを用いて筋が切れても動き続けられるかどうかを表す指標を計算する.
  $G$の$i$行目, つまり$i$番目の筋が切れ, そのモーメントアームが全て0となったものを$G_i$とする.
  本研究では以下の値$E$を評価値として用いる.
  \begin{align}
    r_0 &= \textrm{CalcRITS}(G, \bm{\tau}_{g})\\
    r_i &= \textrm{CalcRITS}(G_i, \bm{\tau}_{g})\;\;\;(1 \leq i \leq M)\\
    E_{value} &= \Sigma^{M}_{i=0}{r_i}\\
    E_{count} &= \Sigma^{M}_{i=0}{\textrm{Integer}(r_i > 0)}\\
    E &= \begin{cases}
      E_{value}\;\;\;(E_{count} \geq M_{min}+1)\\
      0\;\;\;\;\;\;(otherwise)
    \end{cases}
  \end{align}
  筋が一本だけ切れるとしたとき, どの筋を切っても大丈夫かを表す数の下限
  ここで, $M_{min}$は切れても大丈夫な筋の数の下限を表し, $\textrm{Integer}(x)$は$x$が真であれば1, 偽であれば0を返す関数である.
  つまり$E_{count}-1$は切れても大丈夫な筋の個数を表し, これが指定した$M_{min}$より多ければ, $G$とそれぞれの筋を切ったときの$G_i$に関するRITSの合計値が$E$となる.
}%

\subsection{Design Optimization and Detailed Implementation}
\switchlanguage%
{%
  Based on the obtained $E$, we optimize the design ($G$) so that even if one muscle is broken, the remaining muscles can still move the joint.
  Although there are various ways to do this, we apply a genetic algorithm (GA) in this study.
  In this section, we explain the experimental setup including concrete values.
  Using the library \cite{fortin2012deap}, we perform crossbreeding with the function cxBlend with 50\% probability and mutation with the function mutGaussian with 20\% probability.
  The individuals are selected by the function selTournament and the tournament size is set to 3.
  The number of individuals is 200 and the number of generations is 50.
  The minimum and maximum values of muscle Jacobian are set to -0.1 and 0.1 [m], and the values are cropped within the range after each crossing or mutation.
  For simplicity, we limit the calculation of $G$ to two decimal places.
  We set $F^{min} = 0$ [N] and $F^{max} = 200$ [N] for the muscle tension.
  In this study, the experimental results are discussed while changing the number of DOFs $N$, the number of muscles $M$, the lower limit of the number of muscles that can be broken $M_{min}$, and the torque $\bm{\tau}_{g}$ to be generated against gravity, friction, or external force.
}%
{%
  得られた評価値$E$をもとに, 筋が一本切れても残りの筋により関節を動かすことができるよう, 設計($G$)を最適化する.
  これには様々な方法が考えられるが, 本研究では遺伝的アルゴリズムを用いた.
  本節では具体的な数値も含め実験セットアップを説明する.
  ライブラリとして\cite{fortin2012deap}を用い, 50\%の確率で関数cxBlendにより交叉, 20\%の確率で関数mutGaussianにより突然変異を行う.
  個体選択は関数selTournamentで行い, トーナメントサイズを3とする.
  個体数は200, 世代数は50とする.
  また, 筋長ヤコビアンについてはそれぞれの値の最小値と最大値を-0.1, 0.1 [m]とし, 交叉や突然変異の度に範囲内に収まるようにcropを行う.
  また, わかりやすいよう小数点以下は2桁までに制限して計算を行う.
  筋張力については$F^{min} = 0$ [N], $F^{max} = 200$ [N]とした.
  本研究では, 関節の自由度数$N$, 筋数$M$, 切断可能な筋数の下限$M_{min}$, 重力や摩擦等に抗って出すべきトルク$\bm{\tau}_{g}$を変化させながら結果を考察する.
}%

\begin{table}[htb]
  \centering
  \caption{The optimized designs of 1-DOF tendon robot. The values show $10G^T$.}
  {
  \setlength{\arraycolsep}{2pt}
  \setlength{\tabcolsep}{3pt}
  \begin{tabular}{l|ccc}
    & $\tau_g=-5$ & $\tau_g=0$ \\ \hline
    $M=3$ & $\begin{pmatrix}-1&1&1\end{pmatrix}$ & No Solution \\
    $M=4$ & $\begin{pmatrix}-1&1&1&1\end{pmatrix}$ & $\begin{pmatrix}-1&-1&1&1\end{pmatrix}$\\
    $M=5$ & $\begin{pmatrix}-1&-1&1&1&1\end{pmatrix}$ & $\begin{pmatrix}-1&-1&1&1&1\end{pmatrix}$ \\
    $M=6$ & $\begin{pmatrix}-1&-1&-1&1&1&1\end{pmatrix}$ & $\begin{pmatrix}-1&-1&-1&1&1&1\end{pmatrix}$
  \end{tabular}
}
  \label{table:1-dof-sim}
  \vspace{-1.0ex}
\end{table}

\section{Experiments} \label{sec:experiments}
\subsection{1-DOF Simulation} \label{subsec:1-dof-sim}
\begin{figure}[t]
  \centering
  \includegraphics[width=1.0\columnwidth]{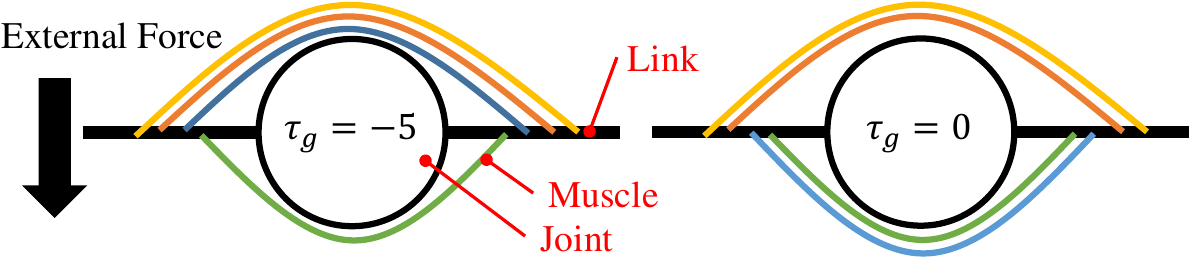}
  \vspace{-3.0ex}
  \caption{The optimized design when $M=4$ and $\tau_g=\{-5, 0\}$.}
  \label{figure:1-dof-sim}
  \vspace{-3.0ex}
\end{figure}

\switchlanguage%
{%
  Simulation experiments are performed on a robot with $N=1$.
  The results of the optimization when the parameters are changed to $M = \{3, 4, 5, 6\}$ and $\tau_{g}=\{0, -5\}$ [Nm] are shown in \tabref{table:1-dof-sim}.
  For ease of viewing, we show $10G^{T}$.
  $M_{min}$ is kept constant with $M_{min}=M$ because the result is the same when $M_{min}$ is changed except when $\tau_g=0$ and $M=3$.

  First, in the case of $M=3$, we cannot get a solution.
  This is because there is only one muscle with positive or negative moment arm, and if it breaks, no torque can be generated in one direction.
  For $M\geq4$, if $M$ is even, the best design is one where the numbers of muscles with positive and negative moment arms are the same.

  On the other hand, when a constant torque is applied like $\tau_g=-5$ (e.g., the elbow bent to -90 [deg]), the solution is obtained even for $M=3$.
  This is because even if the muscle with negative moment arm breaks, the torque can still be maintained by $\tau_g$.
  In the case of $M=4$, the optimal designs for $\tau_g=\{-5, 0\}$ are shown in \figref{figure:1-dof-sim}.
  For the case of $\tau_g=-5$, there are three muscles with positive moment arm and one muscle with negative moment arm, while for $\tau_g=0$, there are two muscles with positive moment arm and two muscles with negative moment arm.
  This means that when $\tau_g=-5$, the positive torque is generated ($\tau_{g}$ is the joint torque that must be generated against gravity), and the available torque space is greater if the number of muscles with positive moment arm is greater than the number of muscles with negative moment arm.
}%
{%
  $N=1$のロボットについてシミュレーションを行う.
  パラメータを$M = \{3, 4, 5, 6\}$, $\tau_{g}=\{0, -5\}$に変化させたときに最適化された結果を\tabref{table:1-dof-sim}に示す.
  なお, 見やすさのため, 10倍した$10G^{T}$を表示している.
  $M_{min}$は変化させても$\tau_g=0, M=3$の時以外は同じ結果であったため, $M_{min}=M$と一定としている.

  まず, $\tau_g=0$の場合については, $M=3$では解が得られていない.
  これは, 必ず正か負のモーメントアームの筋が1本だけになってしまうため, それが破断してしまうと片側に対してトルクが発生できなくなってしまうからである.
  また, $M\geq4$のときは, $M$が偶数であれば正と負のモーメントアームの数が一致している設計が最適であることがわかる.

  これに対して, $\tau_g=-5$のように一定のトルクがかかる(例えば肘を-90度に曲げた)状態では, $M=3$のときも解が出ている.
  これは, もし負のモーメントアームの筋が切れたとしても$\tau_g$によってトルクを確保できるためである.
  また, $M=4$のときについてはわかりやすいように\figref{figure:1-dof-sim}に$\tau_g=\{-5, 0\}$のときの最適設計を示す.
  $\tau_g=-5$の場合には正のモーメントアームの筋が3本, 負が1本であり, $\tau_g=0$のときは両者とも2本であった.
  これは, $\tau_g=-5$の場合は$\tau_g$により正のトルクが稼げるため, 負のモーメントアームより正のモーメントアームの筋を多くしたほうがより発揮可能トルク空間が稼げるということである.
  正のモーメントアームの筋を切ると, $M=3$の場合の最適設計と同じになる点も興味深い.
}%

\begin{table}[htb]
  \centering
  \caption{The optimized design of 2-DOF tendon robot. The values show $10G^T$.}
  {
  \setlength{\arraycolsep}{2pt}
  \setlength{\tabcolsep}{3pt}
  \begin{tabular}{l|ccc}
    & $\bm{\tau}^T_g=\begin{pmatrix}-5&0\end{pmatrix}$ & $\bm{\tau}^T_g=\begin{pmatrix}0&0\end{pmatrix}$ \\ \hline
    \begin{tabular}{c}$M=4$\\$(M_{min}=4)$\end{tabular} & $\begin{pmatrix}-0.5&-0.5&1&1\\-1&1&-0.2&0.2\end{pmatrix}$ & No Solution \\
    \begin{tabular}{c}$M=4$\\$(M_{min}=3)$\end{tabular} & $\begin{pmatrix}-1&0.7&0.8&0.8\\0&1&-1&1\end{pmatrix}$ & No Solution \\
    \begin{tabular}{c}$M=5$\\$(M_{min}=5)$\end{tabular} & $\begin{pmatrix}-1&0.2&0.6&1&1\\-0.4&1&1&-1&-0.9\end{pmatrix}$ & $\begin{pmatrix}-1&-0.8&0&0.8&1\\-0.5&1&-1&1&0.5\end{pmatrix}$ \\
    \begin{tabular}{c}$M=5$\\$(M_{min}=4)$\end{tabular} & $\begin{pmatrix}-1&-1&0.7&0.8&0.9\\0&0.1&-1&1&-1\end{pmatrix}$ & $\begin{pmatrix}-1&-1&0.5&1&1\\-0.4&0.4&-1&-1&1\end{pmatrix}$ \\
  \end{tabular}
  }
  \label{table:2-dof-sim}
  \vspace{-3.0ex}
\end{table}

\begin{figure*}[t]
  \centering
  \includegraphics[width=2.0\columnwidth]{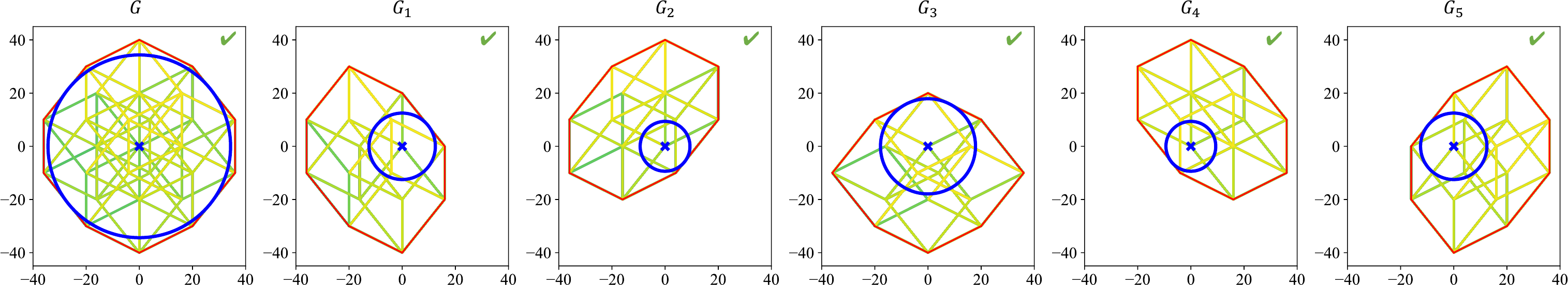}
  \vspace{-1.0ex}
  \caption{The available torque space and RITS of the optimized design when $\bm{\tau}^T_g=\begin{pmatrix}0&0\end{pmatrix}$, $M=5$, and $M_{min}=5$.}
  \label{figure:2-dof-sim-1}
  \vspace{-1.0ex}
\end{figure*}

\begin{figure*}[t]
  \centering
  \includegraphics[width=2.0\columnwidth]{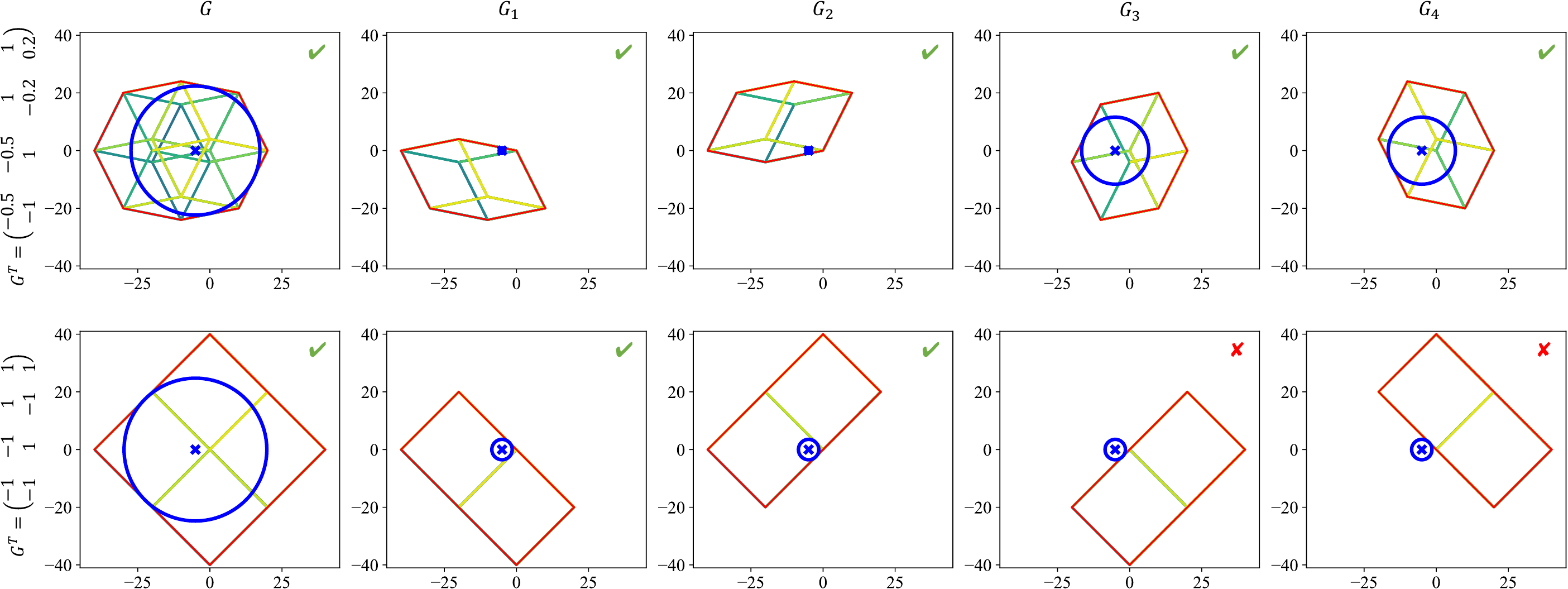}
  \vspace{-1.0ex}
  \caption{The available torque space and RITS of the optimized design when $\bm{\tau}^T_g=\begin{pmatrix}-5&0\end{pmatrix}$, $M=4$, and $M_{min}=4$, and when absolute value of each moment arm of A is maximized with the same sign.}
  \label{figure:2-dof-sim-2}
  \vspace{-1.0ex}
\end{figure*}

\begin{figure*}[t]
  \centering
  \includegraphics[width=1.9\columnwidth]{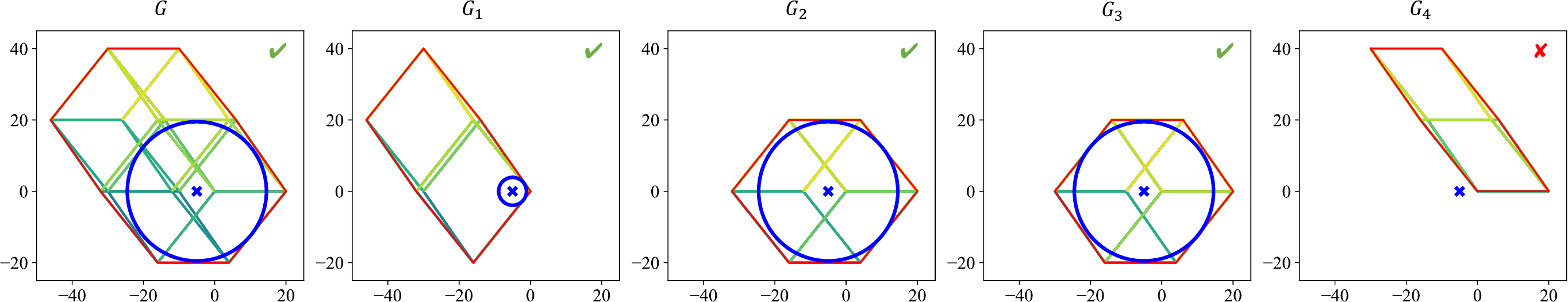}
  \vspace{-1.0ex}
  \caption{The available torque space and RITS of the optimized design when $\bm{\tau}^T_g=\begin{pmatrix}-5&0\end{pmatrix}$, $M=4$, and $M_{min}=3$.}
  \label{figure:2-dof-sim-3}
  \vspace{-3.0ex}
\end{figure*}

\subsection{2-DOF Simulation} \label{subsec:2-dof-sim}
\switchlanguage%
{%
  Simulation experiments are performed on a robot with $N=2$.
  The results of the optimization when the parameters are changed to $M = \{4, 5\}$, $\bm{\tau}_{g}= \{\begin{pmatrix}-5&0\end{pmatrix}^T, \begin{pmatrix}0&0\end{pmatrix}^T\}$, and $M_{min}=\{M, M-1\}$ are shown in \tabref{table:2-dof-sim}.
  For ease of viewing, we show $10G^{T}$.

  First, in the case of $\bm{\tau}^T_g=\begin{pmatrix}0&0\end{pmatrix}$, a solution cannot be found for $M=4$ and $M_{min}=\{4, 3\}$.
  Therefore, in order to move 2 DOFs and to make the robot continue to move when one of the muscles breaks, at least five muscles are necessary.
  Then, for $M=5$ and $M_{min}=5$, the available torque space and RITS of the optimal design, and their changes when one muscle breaks are shown in \figref{figure:2-dof-sim-1}.
  In the optimal design, a large circle is drawn around the origin $\bm{\tau}_{g}$.
  It is also shown that the $\bm{\tau}_{g}$ is included in the available torque spaces when muscles are broken one by one, and the joint can be moved in any direction.

  Second, in the case of $\bm{\tau}^T_g=\begin{pmatrix}-5&0\end{pmatrix}$, a solution is obtained even for $M=4$.
  In contrast to the 1-DOF case, it should be noted that not all moment arms are 1 or -1, but they include values of -0.5 -0.2, and 0.2.
  The torque space and RITS of the optimal design (A) for $M=4$ and $M_{min}=4$, and their changes when the muscles are broken one by one are shown in the upper figure of \figref{figure:2-dof-sim-2}.
  The design (A') when absolute value of each moment arm of A is maximized with the same sign, is also shown in the lower figure of \figref{figure:2-dof-sim-2}.
  The check mark at the top right corner of each figure shows whether $\bm{\tau}_g$ exists in the available torque space or not.
  In the design of A, no matter which muscle is broken, the $\bm{\tau}_g$ remains in the available torque space and the robot can keep moving.
  Note that when muscles 1 and 2 break, the RITS is very small, so in order to solve this problem, we can put a restriction on the size of RITS during optimization.
  However, when the maximum moment arm is distributed evenly to all the joints, as with the design A', when muscles 3 and 4 break, the torque cannot be generated in some directions.
  In other words, it is necessary to set small values of the moment arm depending on the joint, instead of setting them all at the maximum.

  Finally, the results for $\bm{\tau}^T_g=\begin{pmatrix}-5&0\end{pmatrix}$, $M=4$, and $M_{min}=3$ are shown in \figref{figure:2-dof-sim-3}.
  Not limited to these parameters, for $M_{min}=M-1$, the available torque space of the optimized design is likely to be extended in one direction.
  For the muscles 2 and 3, the RITS does not change when each of the muscles is broken.
  Also, since $M_{min}=M-1$, for one certain muscle, the robot cannot move when it is broken.
}%
{%
  $N=2$のロボットについてシミュレーションを行う.
  パラメータを$M = \{4, 5\}$, $\bm{\tau}_{g}= \{\begin{pmatrix}-5&0\end{pmatrix}^T, \begin{pmatrix}0&0\end{pmatrix}^T\}$, $M_{min}=\{M, M-1\}$に変化させたときに最適化された結果を\tabref{table:2-dof-sim}に示す.
  なお, 同様に見やすさのため, 10倍した$10G^{T}$を表示している.

  まず, $\bm{\tau}^T_g=\begin{pmatrix}0&0\end{pmatrix}$の場合は$M=4$のときは$M_{min}=\{4, 3\}$について解が見つからない.
  そのため, 2自由度動かし, かつどの筋が一本切れても大丈夫という状態にするためには, 必ず5本以上の筋が必要となる.
  そして, $M=5, M_{min}=5$のときの最適設計の発揮可能トルク空間とRITS, 筋を1本ずつ切ったときのそれらの変化を\figref{figure:2-dof-sim-1}に示す.
  最適設計では原点$\bm{\tau}_{g}$を中心に大きな円が描かれていることがわかる.
  また, 一本ずつ筋を切った際の$\bm{\tau}_{g}$は発揮可能トルク空間内に内包され, 関節を任意の方向に動かすことができることがわかる.

  次に, $\bm{\tau}^T_g=\begin{pmatrix}-5&0\end{pmatrix}$の場合は, $M=4$についても解が得られている.
  1自由度のときとは違い, 全部のモーメントアームが1か-1ではなく, -0.5や0.2等の値も含まれていることにも着目したい.
  $M=4, M_{min}=4$のときの最適設計(A)の発揮可能トルク空間とRITS, 筋を1本ずつ切ったときのそれらの変化を\figref{figure:2-dof-sim-2}の上段に示す.
  また, -0.5や0.2を-1, 1のように, 符号をそのままに絶対値を最大にした場合の設計(A')についても\figref{figure:2-dof-sim-2}の下段に示す.
  なお, それぞれの図の右上にあるチェックマークは, 発揮可能トルク空間内に$\bm{\tau}_g$が存在するかどうかを表している.
  Aの設計では, どの筋が切れても$\bm{\tau}_{g}$が発揮可能トルク空間内に存在し, 動き続けることができる.
  なお, 筋1と2が切れた際はRITSが非常に小さいため, これを解消するためには, 最適化の際にRITSの大きさにも制約をかけることが考えられる.
  しかし, 通常考えられる, 最大のモーメントアームをどの関節に対しても均等に配置する設計A'では, 筋3と4が切れた際には任意の方向にトルクが出せなくなってしまうことがわかる.
  つまり, モーメントアームは全部最大に設定するのではなく, 場所に応じて小さな値を設定する必要があることがわかる.

  最後に$\bm{\tau}^T_g=\begin{pmatrix}-5&0\end{pmatrix}$, $M=4$, $M_{min}=3$のときの結果を\figref{figure:2-dof-sim-3}に示す.
  これに限らないが, $M_{min}=M-1$のときは, 最適化された設計の発揮可能トルク空間がある一方向に伸びた形になっていることがわかる.
  そして, 2つの筋2と4については, 筋を切ってもRITSが変化しない.
  また, $M_{min}=M-1$であるため, ある一つの筋については切れると完全に動かせなくなってしまうということもわかる.
}%

\begin{figure}[t]
  \centering
  \includegraphics[width=0.7\columnwidth]{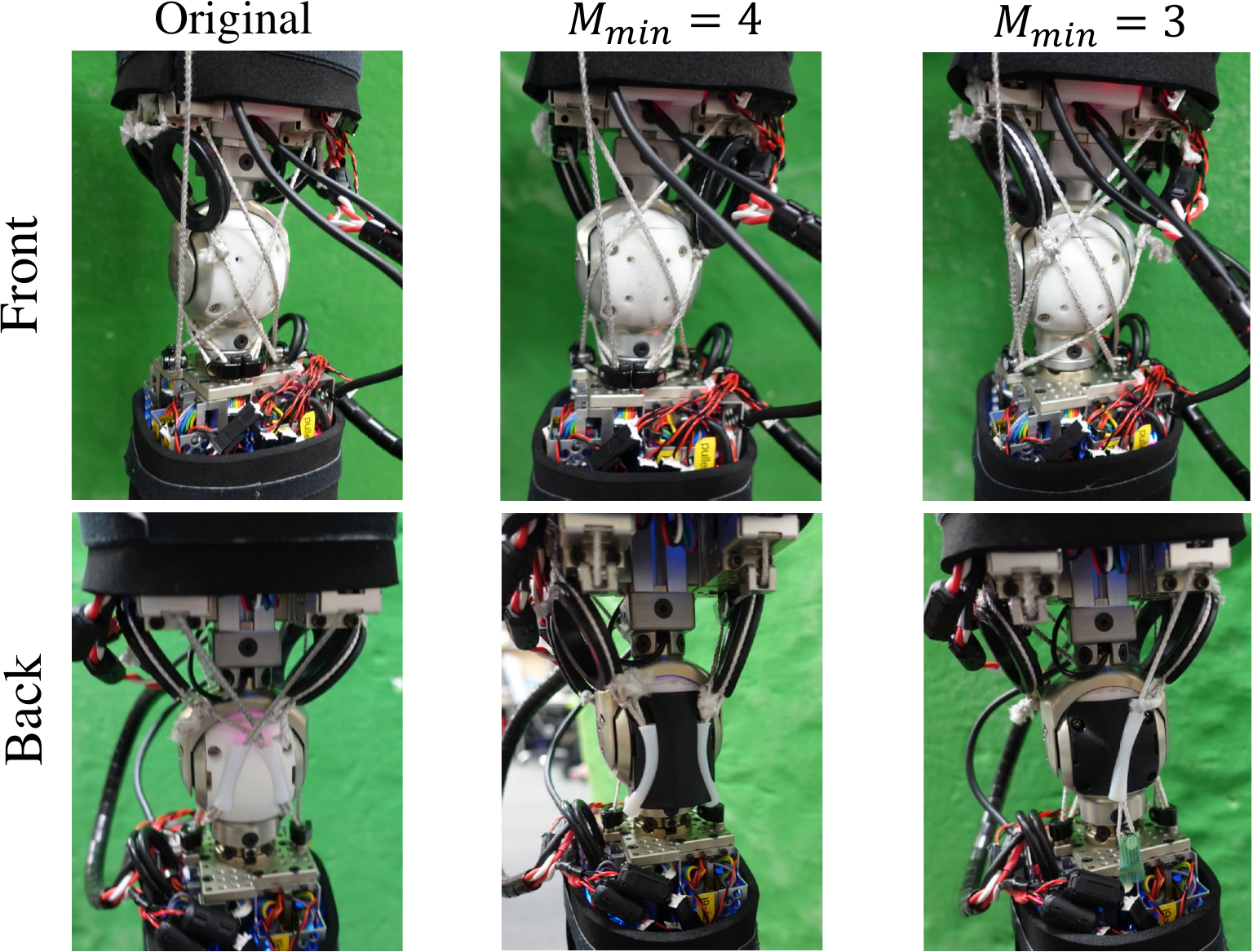}
  \vspace{-1.0ex}
  \caption{The original design and optimized designs with $M_{min}=\{4, 3\}$ of the elbow of Musashi.}
  \label{figure:musashi-muscle}
  \vspace{-3.0ex}
\end{figure}

\begin{table}[htb]
  \centering
  \caption{The human-made designs of the elbow of Musashi with reference to the optimal designs obtained in simulation. The values show $10G^T$.}
  \begin{tabular}{c|ccc}
    original & $\begin{pmatrix}-0.47&-0.26&0.43&0.42\\-0.17&0.20&-0.15&0.24\end{pmatrix}$ \\
    $M=4, M_{min}=4$ & $\begin{pmatrix}-0.18&-0.29&0.61&0.43\\-0.10&0.20&-0.043&0.036\end{pmatrix}$ \\
    $M=4, M_{min}=3$ & $\begin{pmatrix}-0.47&0.50&0.66&0.73\\0.047&-0.19&-0.30&0.18\end{pmatrix}$ \\
  \end{tabular}
  \label{table:musashi}
  \vspace{-3.0ex}
\end{table}

\begin{figure*}[t]
  \centering
  \includegraphics[width=2.0\columnwidth]{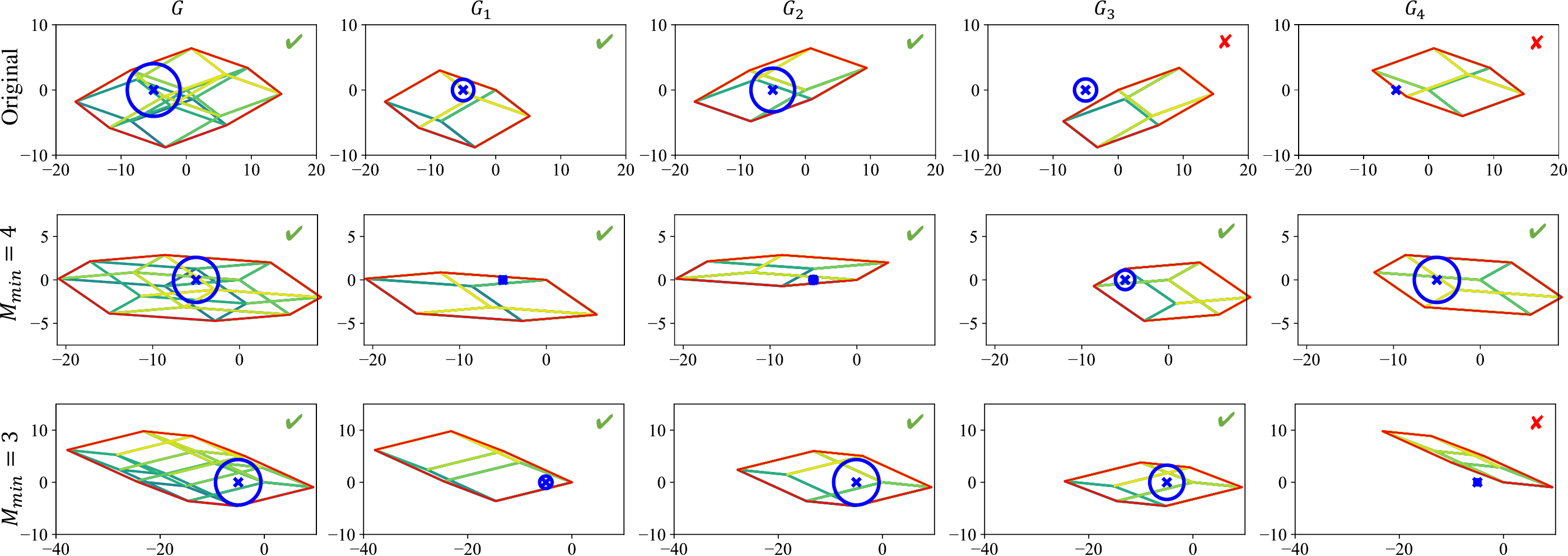}
  \vspace{-1.0ex}
  \caption{The available torque space and RITS of the original design of Musashi and the designs based on the optimization results of $M_{min}=\{4, 3\}$ in \secref{subsec:2-dof-sim}.}
  \label{figure:musashi-torque}
  \vspace{-1.0ex}
\end{figure*}

\begin{figure*}[t]
  \centering
  \includegraphics[width=1.9\columnwidth]{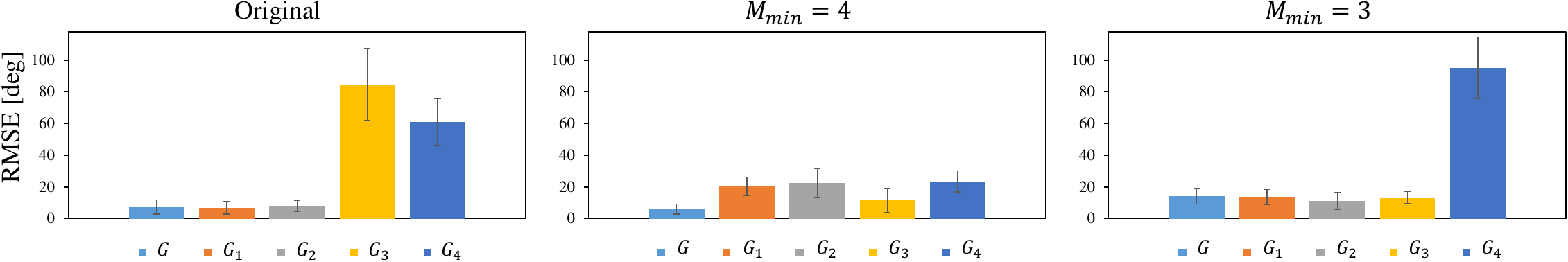}
  \vspace{-1.0ex}
  \caption{The average and standard deviation of RMSE between the commanded and measured joint angles when using the original design and the optimized designs with $M_{min}=\{4, 3\}$.}
  \label{figure:musashi-eval}
  \vspace{-3.0ex}
\end{figure*}

\subsection{Actual Musculoskeletal Humanoid}
\switchlanguage%
{%
  For the simulation of $N=2$ in \secref{subsec:2-dof-sim}, we consider the state of $M=4$ and $\bm{\tau}^T_g=\begin{pmatrix}-5&0\end{pmatrix}$.
  The left elbow of the musculoskeletal humanoid Musashi \cite{kawaharazuka2019musashi} has four muscles and two joints, elbow-p and elbow-y (-p and -y means pitch and yaw, respectively).
  When the elbow is bent about -60 [deg] (the posture is expressed as $\bm{\theta}_{0}$), a torque of about $\begin{pmatrix}-5&0\end{pmatrix}$ is required for gravity compensation, which is similar to the simulation condition in \secref{subsec:2-dof-sim} ($\bm{\tau}^T_g=\begin{pmatrix}\tau_{elbow-p}&\tau_{elbow-y}\end{pmatrix}$).
  Therefore, we modify the muscle arrangement around the elbow with reference to $G$ obtained from the simulation results to verify the difference in performance on the actual robot Musashi.
  Note that the muscles of Musashi do not have precise constant moment arms with pulleys, but are wrapped around the ball joints like a human being, so the design is somewhat dependent on the human design and interpretation.

  The designed muscle arrangements are shown in \figref{figure:musashi-muscle}.
  From left to right, they are the original design of Musashi \cite{kawaharazuka2019musashi} and the designs based on the optimization results of $M_{min}=4$ and $M_{min}=3$.
  Note that the original one is similar to the design in the lower part of \figref{figure:2-dof-sim-2}.

  Since the moment arm cannot be seen from the figure, the relationship between joint angle, muscle tension, and muscle length is learned using the method of \cite{kawaharazuka2020autoencoder}, and the obtained muscle Jacobian around $\bm{\theta}_{0}$ is shown in \tabref{table:musashi}.
  For ease of viewing, we show $10G^T$.

  While the original design has similar moment arms for each elbow-p and elbow-y, regarding $M_{min}=4$, some moment arms are made small to reduce the interference between muscles.
  Regarding $M_{min}=3$, three muscles with positive moment arms are used for elbow-p instead of two muscles with positive moment arm and two muscles with negative moment arm.
  The available torque space and RITS obtained from the moment arm $G$ and their changes when muscles are broken one by one are shown in \figref{figure:musashi-torque}.
  The available torque space in the direction of elbow-y (y-axis of the figure) is shrunk more than that of \figref{figure:2-dof-sim-2} and \figref{figure:2-dof-sim-3}, because the moment arm for elbow-y is difficult to obtain due to design reasons.
  The upper part of \figref{figure:musashi-torque} and the lower part of \figref{figure:2-dof-sim-2}, the middle part of \figref{figure:musashi-torque} and the upper part of \figref{figure:2-dof-sim-2}, and the lower part of \figref{figure:musashi-torque} and \figref{figure:2-dof-sim-3} correspond to each other, and the shape of their available torque space, RITS, and the ability to generate force in any direction are similar to each other.

  With these obtained muscle arrangements, six postures of elbow-p and elbow-y around $\bm{\theta}_{0}$ are determined, and the control input is sent to the actual robot using a controller of \cite{kawaharazuka2020autoencoder}, which is repeated three times.
  For each of the designs, this experiment is performed for a design with no muscle rupture and for a design with a single muscle rupture (with the motor current set to zero), and the mean and variance of the RMSE (Root Mean Squared Error) between the commanded and measured joint angles are calculated.
  The results are shown in \figref{figure:musashi-eval}.

  In the original design, the RMSE rises so much at the ruptures of the third and fourth muscles that the joints do not move well, and the result is the same as the upper part of \figref{figure:musashi-torque}.
  For the optimized design with $M_{min}=4$, the RMSE increases slightly by each muscle rupture, but the body motion is successfully maintained regardless of which muscle is broken.
  The RMSE regarding the optimized design with $M_{min}=3$ increases significantly when the fourth muscle is broken, and the result is the same as the lower part of \figref{figure:musashi-torque}.
  Therefore, we found that this method is also effective in the actual robot.
}%
{%
  $N=2$のシミュレーションについて, \secref{subsec:2-dof-sim}の$M=4$, $\bm{\tau}^T_g=\begin{pmatrix}-5&0\end{pmatrix}$の状態を考える.
  筋骨格ヒューマノイドMusashiの左肘には4つの筋とelbow-pitch, elbow-yawの2つの関節が存在する.
  そのため, 肘を-60度程度曲げたとき($\bm{\theta}_{0}$)の状態は, 大体$\bm{\tau}^T_g=\begin{pmatrix}-5&0\end{pmatrix}$の力が重力補償に必要であり, 前節の最適化の条件と似た状態になっている.
  そこで, シミュレーションから得られた$G$を参考に, 肘周りの筋経路を修正し, 実機においてどのような性能差が生まれるかについて検証する.
  筋骨格ヒューマノイドMusashiは筋経路がプーリ等により正確に設計できるわけではなく, 人間と同じように球関節に巻きつく形であるため, 人間の設計・解釈性に多少依存することに留意されたい.

  設計した筋経路を\figref{figure:musashi-muscle}に示す.
  左から, これまで使われてきた筋経路\cite{kawaharazuka2019musashi}, $M_{min}=4$の最適設計解, $M_{min}=3$の最適設計解を参考にしたものである.
  なお, このOriginalは\figref{figure:2-dof-sim-2}の下段の設計に似通っている.
  しかし, 図からではモーメントアームはわからないため, \cite{kawaharazuka2020autoencoder}の手法を用いて関節・筋張力・筋長の関係を学習し, 得られた$\bm{\theta}_{0}$における筋長ヤコビアンを\tabref{table:musashi}に示す.
  なお, 見やすさから同様に$10G^T$を表示している.
  Originalはelbow-p, elbow-yについてそれぞれの筋が同じようなモーメントアームを有しているのに対して, $M_{min}=4$の設計解を参考にした筋配置では一部のモーメントアームを小さく設定することで筋張力の干渉を減らしている.
  また, $M_{min}=3$の設計解を参考にした筋配置では, elbow-pに対する正負のモーメントアームを持つ筋を2つず配置するのではなく, 正のモーメントアームを持つ筋を3つにして偏りを持たせている.
  このときのモーメントアーム$G$から得られる発揮可能トルク空間とRITS, 筋を1本ずつ切ったときのそれらの変化を\figref{figure:musashi-torque}に示す.
  elbow-yaw (図のy軸)に関するモーメントアームが設計の都合上得にくいため, \figref{figure:2-dof-sim-2}や\figref{figure:2-dof-sim-3}よりも縦方向が縮んでいる.
  \figref{figure:musashi-torque}の上段と\figref{figure:2-dof-sim-2}の下段, \figref{figure:musashi-torque}の中段と\figref{figure:2-dof-sim-2}の上段, \figref{figure:musashi-torque}の下段と\figref{figure:2-dof-sim-3}が対応しており, それらの発揮可能トルク空間の形, RITS, 任意方向に力を発揮可能かどうかが似通っていることが読み取れる.

  これら得られた筋配置の状態で, $\bm{\theta}_{0}$周辺のelbow-p, elbow-yの関節角度を6つ決め, 実機に\cite{kawaharazuka2020autoencoder}を用いて制御入力として送ることを3回繰り返す.
  それぞれの設計について, 筋が一本も切れていない状態と, それぞれの筋が一本ずつ切れた状態(電流を0とした)についてこの実験を行い, 指令関節角度と測定された関節角度のRMSE (Root Mean Squared Error)の平均と分散を計算する.
  その結果を\figref{figure:musashi-eval}に示す.
  Originalでは3本目, 4本目を切った際にRMSEが大きく上昇し上手く身体を動かせておらず, \figref{figure:musashi-torque}の上段と同じ結果となった.
  $M_{min}=4$で最適化された設計については, 筋を切ることで少しRMSEは上昇するものの, どの筋を切っても身体の運動を維持することに成功した.
  $M_{min}=3$で最適化された設計については, 4本目の筋を切った際はRMSEが大きく上昇し上手く身体を動かせておらず, \figref{figure:musashi-torque}の下段と同じ結果となった.
  ゆえに, 実機においても本手法が有効であることがわかった.
}%

\section{Discussion} \label{sec:discussion}
\switchlanguage%
{%
  We discuss the results obtained from the experiments of this study.
  First, experiments on the 1-DOF simulation show that the number of muscles with positive and negative moment arms should be as equal as possible in order to maximize the effect of redundancy for the 1-DOF robot with $\tau_g=0$.
  In this case, the moment arms of all muscles should be the maximum value.
  If extra force needs to be applied in one direction, as in $\tau_g=-5$, by increasing the number of muscles in that direction, the RITS can be large even if one of the muscles is broken.

  Next, experiments on the 2-DOF simulation show that in the case with $\bm{\tau}^T_g=\begin{pmatrix}0&0\end{pmatrix}$, at least five muscles are necessary to be able to move the joint in any direction even if one muscle breaks.
  This result is consistent with the analysis in \cite{valero2015neuromechanics} that at least $2N$ muscles are required to move $N$ DOFs versatilely.
  If extra force needs to be applied in one direction, such as in the case of $\bm{\tau}^T_g=\begin{pmatrix}-5&0\end{pmatrix}$, four muscles may be enough, but the RITS when some muscles are broken may be very small.
  Also, unlike the case of the 1-DOF robot, if the absolute value of each moment arm is set to the maximum, the joint will not be able to make use of the redundancy because the balance of muscle tensions will not be maintained when one muscle is broken.
  Therefore, in order to take advantage of the redundancy, it is important to reduce moment arms of some muscles to half or less.
  Also, if we relax the constraint of $M_{min}$ like $M_{min}=M-1$, we can derive a design in which one of the muscles plays an important role and RITS can be kept large when the rest of the muscles is broken.
  Since only the important muscle is required to be protected, we can consider taking advantage of the redundancy by actively using such a design.

  Finally, from the actual robot experiments of Musashi, it is shown that the same performance can be obtained for the actual robot by imitating the optimal design obtained in simulations.
  Although the difference between the commanded and the realized joint angles may be larger when the RITS is small due to the influence of friction, which is difficult to modelize on the actual robot, the characteristics of the actual robot are almost the same with the simulation results.
  Therefore, it is expected that the muscle arrangement can be modified based on the simulation results to create a robust body that can continue to move even if one muscle breaks.

  The limitations of this study are described below.
  First, although this study can be applied to the general system with more than two degrees of freedom, the computational complexity of $E$ depends on calculation of all the vertices of the hypercube $F$ in the RITS calculation O($2^M$), and therefore, it explodes exponentially with the number of muscles.
  Since the number of muscles should be increased as the number of joints increases, the limit is currently about $N=3$ and $M=7$, and the speed of the algorithm should be increased.
  Using the conversion between span-form and face-form \cite{fukuda1996facespan} is one of its solutions.
  Second, the optimization is performed based on the approximation of constant moment arm, and it is applied to the design of human-like joints where the moment arm is not constant.
  Therefore, we kept $N=2$ in this study because the more $N$ is increased, the more the design depends on the human interpretation.
  In the future, it is necessary to make a method generally applicable to the system with variable moment arms.
  We also would like to complete a series of processes as in \cite{kwiatkowski2019selfmodeling}, which detects physical changes such as a muscle break, relearns self-models, and adapts to the real world.
}%
{%
  本研究の実験から得られた結果について考察する.
  まず, 1-DOFのシミュレーションに関する実験から, 1自由度ロボットで$\tau_g=0$の場合に冗長性の効果を最大化するには, 正と負のモーメントアームの筋を出来る限り同じ本数にするべきであることがわかった.
  この際, モーメントアームは全ての筋について設定した最大値である方が良い.
  また, $\tau_g=-5$のように一方方向に余分に力をかける必要がある状態においては, その方向に筋を多めに配置することで, どの筋が一本切れてもRITSを大きく取ることができる.

  次に, 2-DOFのシミュレーションに関する実験から, 2自由度ロボットで$\bm{\tau}^T_g=\begin{pmatrix}0&0\end{pmatrix}$の場合には, どの筋が切れても任意方向に関節を動かせるようにするためには, 5本以上の筋が必要であることがわかった.
  $\bm{\tau}^T_g=\begin{pmatrix}-5&0\end{pmatrix}$のように一歩方向に余分に力をかける必要がある状態においては, 4本の筋でも大丈夫であるが, 一部の筋が切れた際のRITSが非常に小さくなってしまう場合がある.
  また, 1自由度の場合と違い, 全モーメントアームの絶対値を同じように最大にしてしまうと一部の筋が切れた時に釣り合いが保てず関節が動かせなくなってしまう.
  そのため, 一部のモーメントアームを半分やそれ以下に小さくすることが冗長性を活かすためには重要となる.
  また, $M_{min}=M-1$のように制約を緩和すると, その一本に大きな役割を担わせ, 残りの筋は切れてもRITSを大きく保てるような設計が導出される.
  これは逆に, その重要な一本さえ守れば良いため, これを積極的に利用した冗長性の利用も考えられうる.

  最後に, Musashiの実機に関する実験から, シミュレーションで得られた最適設計解を模倣することで, 実機に置いても同様の性能が得られることがわかった.
  実機に置いてはモデル化が難しい摩擦の影響等があるため, RITSが小さい場合は指令関節角度と実現された関節角度の差が多少大きくなる場合があるが, 概ね特性はシミュレーションと一致していた.
  よって, シミュレーションで得られた結果を元に設計を変更することで, 筋が一本切れても動き続けるロバストな身体を作ることができると考える.

  本研究の限界について述べる.
  まず, 本研究は関節が3自由度以上の一般の系に対しても適用可能であるが, 評価値$E$の計算量はRITSの計算における超立方体の全頂点列挙の計算量O($2^M$)に依存するため, 筋数に従って指数関数的に爆発する.
  これは, face-span変換を利用しても同じである.
  関節が増えるに従って筋数も増やす必要があるため, 現状$N=3, M=7$程度までが限界であり, 今後高速化の必要がある.
  また, 本研究はモーメントアームを一定と近似したうえで最適化を行い, モーメントアームが一定でない人間のような関節の設計にそれを応用している.
  そのため, $N$が増えるほど人間の解釈に依るところが大きくなってしまうため, 本研究では$N=2$に留めている.
  今後, モーメントアームが$\bm{\theta}$によって徐々に変化する系に対しても一般に適用可能な方法にしていく必要がある.
  また, \cite{kwiatkowski2019selfmodeling}のように, 筋が切れたという身体の変化を検知し, 学習し直して現実世界に適応していくという一連の流れまで行いたい.
}%

\section{Conclusion} \label{sec:conclusion}
\switchlanguage%
{%
  In this study, we proposed a body design analysis and optimization method to take advantage of the redundant muscle arrangement in musculoskeletal humanoids, which allows the body to continue to move even if one muscle breaks.
  We described the method of maximizing the radius of the hypersphere inscribed in the hyperpolyhedron of the available joint torque space (RITS) when the muscles are broken one by one.
  The simulation experiments show that instead of having all moment arms arranged in the same way, it is possible to make some moment arms smaller so that the robot can continue to move even if one muscle is broken.
  Depending on the evaluation function of the optimization, it is also possible to create a biased design where the load is concentrated on one muscle and the robot can move even if any of the other muscles break.
  The application of this study to the actual musculoskeletal humanoid generated the same results as in the simulations, which confirmed the effectiveness of this study.

  In future works, we would like to extend these results to the whole body design to make the robot more robust.
}%
{%
  本研究では, 筋骨格ヒューマノイドの冗長な筋配置の特徴である, 筋が破断しても身体を動かし続けられるという利点を最大限に活かすための解析手法・身体設計最適化について提案した.
  筋が一本ずつ切れた際に発揮可能関節トルク空間が張る超多面体に内接する超球の半径を最大化するという問題を設定し, その手法について述べた.
  シミュレーション実験から, 全モーメントアームを同じように配置するのではなく, 一部のモーメントアームを小さくすることで, どの筋が切れても動き続けられるような設計が可能となる.
  また, 最適化の評価関数次第では, 一本に負荷を集中させ, その他の筋はどれが切れても動ける, というような偏った設計にすることも可能であった.
  本研究を実機に適用した結果, シミュレーションと同様の結果が示され, その有効性が確認された.

  今後は, これを全身へと拡張し, よりロバストなロボットを目指していきたい.
}%

{
  \bibliographystyle{IEEEtran}
  \bibliography{main}
}

\end{document}